\documentclass[a4paper,fleqn]{cas-dc}

\usepackage[authoryear,longnamesfirst]{natbib}

\usepackage{algorithm}
\usepackage{algorithmic}
\usepackage{graphicx}
\usepackage{bm}  
\usepackage{subcaption}
\def\tsc#1{\csdef{#1}{\textsc{\lowercase{#1}}\xspace}}
\tsc{WGM}
\tsc{QE}
\tsc{EP}
\tsc{PMS}
\tsc{BEC}
\tsc{DE}

\begin{document}
\let\WriteBookmarks\relax
\def\floatpagepagefraction{1}
\def\textpagefraction{.001}

\shorttitle{Knowledge-Based Systems}

\shortauthors{CV Radhakrishnan et~al.}

\title [mode = title]{CausalDiffTab: Mixed-Type Causal-Aware Diffusion for Tabular Data Generation}                      

\cortext[1]{Corresponding author}
\author[1]{Jia-Chen Zhang}[orcid=0009-0004-5596-3952]
\ead{m325123603@sues.edu.cn}

\author[1]{Zheng Zhou}[orcid=0009-0009-3003-9305]
\ead{M320123332@sues.edu.cn}

\author[1]{Yu-Jie Xiong}[orcid=0000-0002-2769-022X]
\ead{xiong@sues.edu.cn}
\cormark[1]

\author[1]{Chun-Ming Xia}
\cormark[1]
\ead{cmxia@sues.edu.cn}

\author[2]{Fei Dai}
\ead{fdai@fudan.edu.cn}

\affiliation[1]{organization={School of Electronic and Electrical Engineering, Shanghai University of Engineering Science},
                addressline={\\333 Longteng Road, Putuo District}, 
                city={Shanghai},
                postcode={201620}, 
                state={Shanghai},
                country={China}}
                
\affiliation[2]{organization={Key Laboratory of Computational Neuroscience and Brain-Inspired Intelligence, Fudan University},
                addressline={\\220 Handan Road, Pudong New District}, 
                city={Shanghai},
                postcode={200433}, 
                state={Shanghai},
                country={China}}
\begin{abstract}
Training data has been proven to be one of the most critical components in training generative AI. However, obtaining high-quality data remains challenging, with data privacy issues presenting a significant hurdle. To address the need for high-quality data. Synthesize data has emerged as a mainstream solution, demonstrating impressive performance in areas such as images, audio, and video. Generating mixed-type data, especially high-quality tabular data, still faces significant challenges. These primarily include its inherent heterogeneous data types, complex inter-variable relationships, and intricate column-wise distributions.
In this paper, we introduce CausalDiffTab, a diffusion model-based generative model specifically designed to handle mixed tabular data containing both numerical and categorical features, while being more flexible in capturing complex interactions among variables. We further propose a hybrid adaptive causal regularization method based on the principle of Hierarchical Prior Fusion. This approach adaptively controls the weight of causal regularization, enhancing the model's performance without compromising its generative capabilities. Comprehensive experiments conducted on seven datasets demonstrate that CausalDiffTab outperforms baseline methods across all metrics. Our code is publicly available at: \url{https://github.com/Godz-z/CausalDiffTab}.
\end{abstract}




\maketitle

\section{Introduction}
Generative large language models, as one of the core technologies in the field of artificial intelligence, exhibit tremendous potential in natural language processing \cite{he2023debertav}, content creation \cite{llama2}, and data analysis \cite{Achiam2023GPT4TR}. One of the factors constraining the development of generative large language models is the lack of high-quality datasets \cite{wang2024do}. Currently, a mainstream solution is to utilize diffusion-based generative models to create datasets, a method that has demonstrated impressive results in domains such as images, audio, and video \cite{LENG2024102627, yang2024editworldsimulatingworlddynamics, DBLP:conf/icml/LiuCYMLM0P23}. However, unlike data composed of purely continuous pixel values with local spatial correlations—such as images—tabular data combines numerical fields (e.g., age, income) with categorical fields (e.g., gender, occupation), presenting more complex and diverse data characteristics. Tabular data is prevalent in various databases and constitutes a core component of data processing and analysis tasks in the information technology domain \cite{fonseca2023tabular, you2020handling, zheng2022diffusion}. Constructing efficient tabular data generation models has become an important research topic, with applications ranging from training data-based guidance to data privacy protection and beyond \cite{assefa2020generating, hernandez2022synthetic}. These technologies play a critical role in modern data management and analysis.
\begin{figure}
  \includegraphics[width=\linewidth]{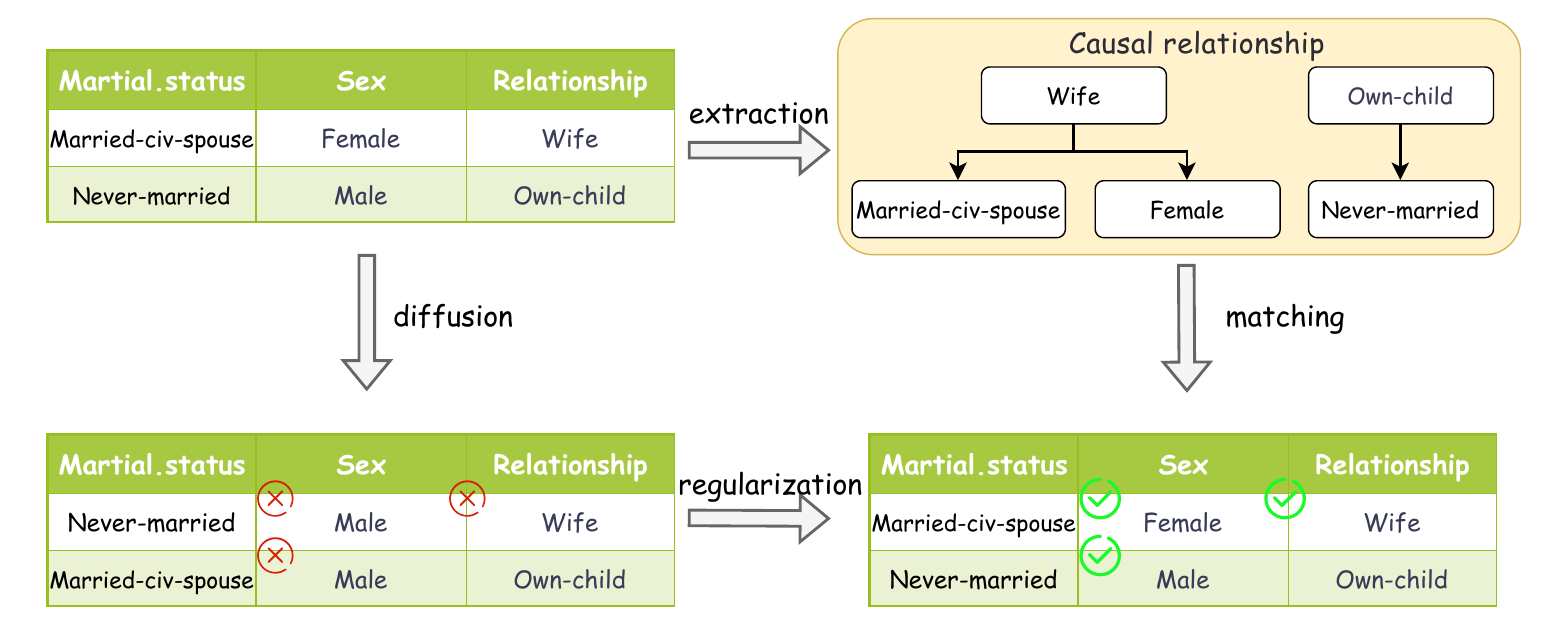}
  \caption{A simple example of causal awareness}
  \vspace{-10pt}
  \label{fig:example}
\end{figure}

\begin{figure}
  \includegraphics[width=\linewidth]{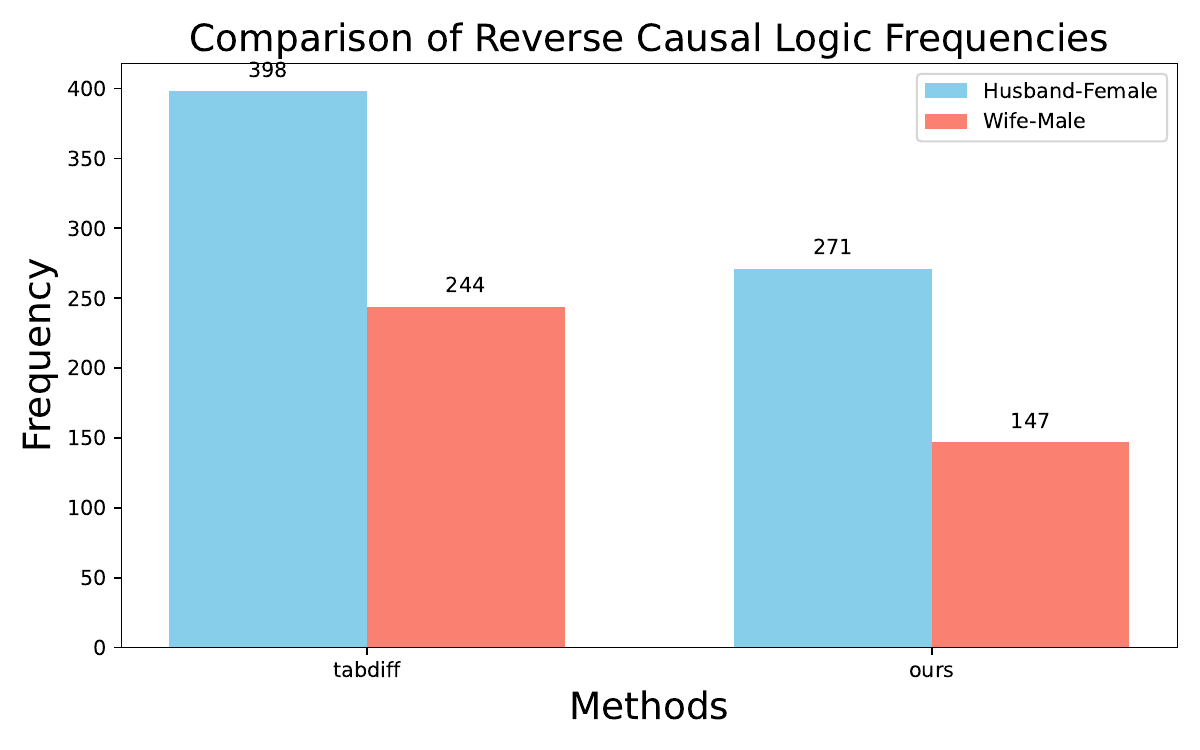}
  \caption{A comparison of whether causal regularization is included. The results show the number of causally implausible instances (Husband-Female and Wife-Male) in the Adult dataset. Our method significantly reduces the number of such causally implausible cases.}
  \vspace{-10pt}
  \label{fig:comparison}
\end{figure}

Traditional statistical models, such as Gaussian mixture models, fit data using simple probability distributions, which limits their expressive capability \cite{borisov2023language}. Variational Autoencoders (VAEs) achieve end-to-end generation by introducing a latent variable encoding-decoding mechanism but are constrained by the blurry quality of their outputs \cite{liu2023goggle}. Subsequently, Generative Adversarial Network (GAN)-based models break through the clarity bottleneck via adversarial training, yet they still face challenges related to training instability \cite{NEURIPS2019_254ed7d2}. In recent years, diffusion models have demonstrated impressive performance in the field of generative modeling \cite{NEURIPS2019_3001ef25, NEURIPS2020_4c5bcfec, Rombach_2022_CVPR}. These models capture complex data distributions through progressive noise perturbation and reverse denoising mechanisms. Researchers have been actively exploring ways to extend this powerful framework to tabular data \cite{10.1145/3534678.3539454, 10.5555/3618408.3619133, zhang2024mixedtype}. However, since diffusion models independently model the conditional probability distributions of individual labels during the generation process. Previous methods struggle to learn causal relationships between different labels, leading to the emergence of counterfactual reasoning phenomena as illustrated in Figure \ref{fig:comparison}.
To address counterfactual issues in generative processes, DiffPO employs a tailored conditional denoising diffusion model to learn complex distributions \cite{ma2024diffpo}. ECI achieves effective label representations by progressively updating event context representations \cite{man2024mastering}. CausalDiffAE enhances the model's causal awareness by mapping high-dimensional data to latent variables with causal relationships through a learnable causal encoder \cite{komanduri2024causal}. CaPaint integrates causal reasoning with generative models to handle counterfactual inference and missing data imputation in spatiotemporal dynamics \cite{duan2024causal}. While these methods significantly strengthen causal awareness capabilities, applying causal awareness to the tabular data generation domain remains challenging due to difficulties in multi-type data awareness and training instability.

Since the causal relationships in tabular data are often highly nonlinear (e.g., interpersonal relationships in the Adult dataset, or conditional causal effects in financial data), linear causal regularization fail to capture such complex relationships due to insufficient modeling capacity. Additionally, the assumption of enforced linear relationships between variables conflicts with the nonlinear architecture of diffusion models \cite{pmlr-v177-uemura22a}. To address these challenges, this paper proposes CausalDiffTab, a novel mixed-type diffusion framework for tabular data generation. The key distinction between CausalDiffTab and existing diffusion-based methods lies in its nonlinear causal modeling of mixed-type data relationships through directed acyclic graph (DAG) construction. Subsequently, the framework aligns causal matrices with noise directions during the diffusion process, enabling more faithful representation of complex causal dependencies. Figure \ref{fig:example} provides an easily understandable example of this process. Furthermore, to prevent excessive causal regularization in the early stages of training from hindering model convergence, we propose a hybrid adaptive causal regularization, which effectively ensures stable training of the model while enhancing robustness to noise during training. We select six representative tasks involving complex tabular data generation from real-world scenarios and adopt seven evaluation metrics to assess the performance of CausalDiffTab in terms of Fidelity, Downstream Task utility, and privacy preservation. All experimental results demonstrate that our method achieves significant performance improvements across multiple scenarios. Our main contributions are as follows:
\begin{itemize}
\item[$\bullet$] We propose a novel complex tabular data generation model called CausalDiffTab, which learns the joint distribution in the original data space through a continuous-time diffusion model.
\end{itemize}
\begin{itemize}
\item[$\bullet$] To ensure the generative capability of the model, we propose hybrid adaptive causal regularization, which effectively enhances the training stability and robustness of the model.
\end{itemize}
\begin{itemize}
\item[$\bullet$]We conduct extensive experiments to validate the effectiveness of our method, performing a comprehensive evaluation on seven datasets across seven metrics. The results demonstrate that CausalDiffTab outperforms the latest baselines in most tasks.
\end{itemize}

\begin{figure*}
  \includegraphics[width=\textwidth]{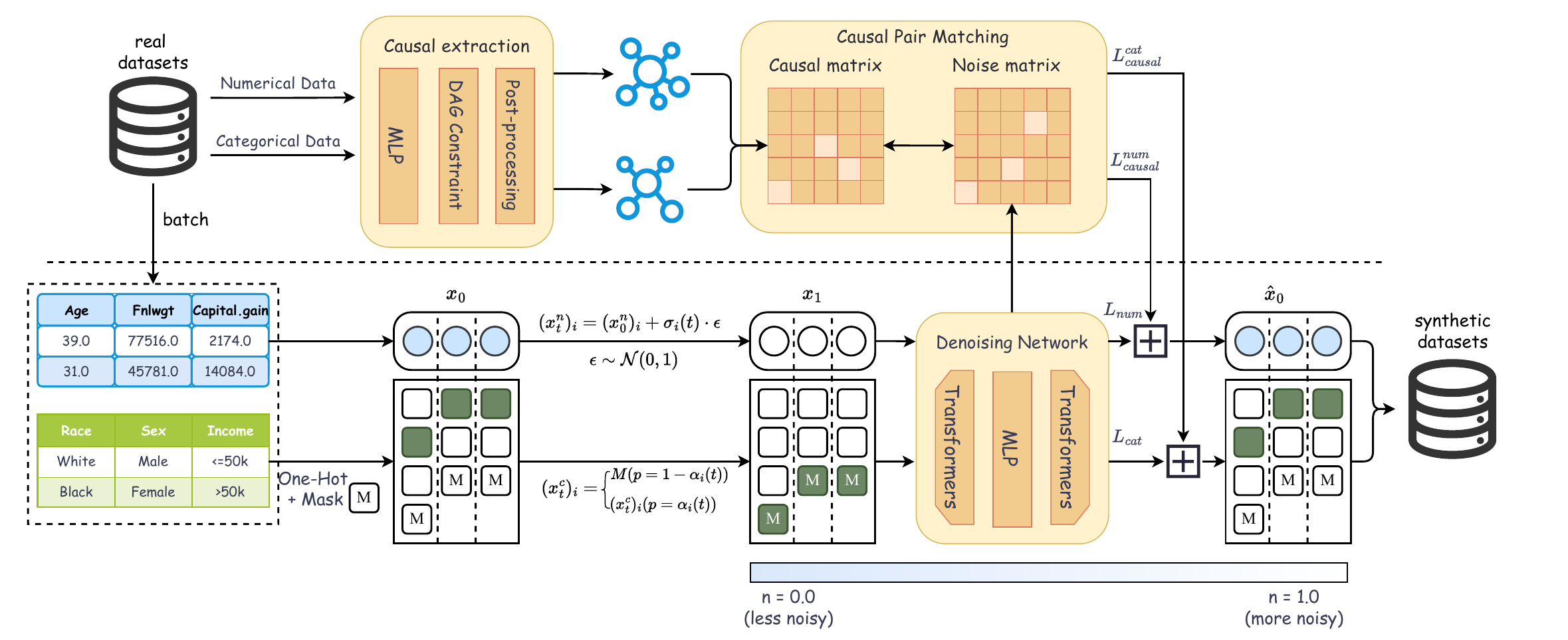}
    \vspace{-10pt}
  \caption{This section provides a high-level overview of CausalDiffTab. The model constructs a causal correlation matrix by applying one-hot encoding to categorical features, thereby establishing interpretable causal relationships between different feature types . During the reverse denoising process, it dynamically aligns this causal matrix with the noise prediction matrix generated at each diffusion step via a causality-constrained mechanism.}
  \label{fig:main}
\end{figure*}
\section{Related Works}
Generative models represent a key research direction in the field of artificial intelligence, generating new samples by learning data distributions and achieving breakthroughs in areas such as image generation and natural language processing \cite{NEURIPS2020_4c5bcfec, devlin-etal-2019-bert, zhang-etal-2025-parameter}. VAEs are based on an encoder-decoder architecture, optimizing the latent space through variational inference \cite{kingma2022autoencodingvariationalbayes}. However, the distribution assumptions limit the model's expressiveness, often resulting in blurry generated samples. Subsequently, GANs were proposed, which fit data distributions through an adversarial game between a generator and a discriminator \cite{10.1145/3422622}. Wasserstein GAN (WGAN) introduced the Wasserstein distance to replace the Jensen-Shannon divergence used in traditional GANs, theoretically alleviating issues like gradient vanishing and training instability \cite{arjovsky2017wassersteingan}. StyleGAN series focuses on improving generation quality and controllability, with GAN-based models being widely applied in fields such as image synthesis and data augmentation \cite{Karras_2019_CVPR, Karras_2020_CVPR}. Some methods also reduce overfitting and increase training stability by combining VAEs with GANs \cite{YAN2025112899}. Nevertheless, challenges like training instability and mode collapse remain difficult to fully resolve. Autoregressive generative models, such as GPT, rely on sequence prediction mechanisms to generate data element by element. While autoregressive models excel in natural language generation, they struggle with efficiency in complex tabular data and image generation, finding it difficult to capture high-dimensional structural dependencies.

In recent years, the powerful generative capabilities of Diffusion models have been impressive \cite{NEURIPS2020_4c5bcfec, DBLP:journals/corr/abs-2311-15127, Meng_2023_CVPR, song2021denoising}. They achieve high-quality sample generation by simulating a forward diffusion process (gradually adding noise to corrupt the data) and a reverse denoising process (learning to recover the original data). The core idea originates from non-equilibrium thermodynamics, establishing a bidirectional mapping between the data distribution and the noise distribution step-by-step through a Markov chain. Building on this foundation, DDIM breaks through the Markov assumption by proposing a non-Markovian sampling strategy \cite{song2021denoising}. By designing an implicit noise transfer function, the number of generation steps is compressed to just a few dozen. Due to the powerful performance of Diffusion models, researchers have begun to focus on how to extend this robust model to table generation. CoDi handles continuous and discrete variables separately using two diffusion models \cite{pmlr-v202-lee23i}. TabCSDI proposes three encoding methods—One-hot, analog bits, and Feature Tokenization—to process input data \cite{zheng2022diffusion}. TabSAL enhances the generation capability of tabular data through a small surrogate auxiliary language model \cite{LI2024112438}. GenerativeMTD generates pseudo-real data from real data to expand the training samples for training deep learning models \cite{SIVAKUMAR2023110956}. Tabsyn synthesizes tabular data by utilizing a diffusion model in the latent space constructed by a VAE \cite{zhang2024mixedtype}. TabDiff employs transformers to handle different input types, building an end-to-end generative framework \cite{shi2025tabdiff}. However, all these models are based on modeling the distribution of the data, while ignoring the causal relationships between the data. Our proposed CausalDiffTab derives a causal matrix from the dataset and then prunes gradients that violate causal directions through causality-constrain, enabling the model to perceive causal relationships among the data and effectively guide the generation process.
\section{Our Methods}
In this section, we first introduce the relevant theories, followed by a detailed description of the methodological details of CausalDiffTab. It is a data-driven, diffusion-based model. The overall architecture is illustrated in Figure \ref{fig:main}.
\subsection{Preliminary}
\subsubsection{Diffusion Model}
Diffusion Model represents a significant breakthrough in the field of generative models in recent years. It models data distributions through a gradual denoising mechanism, with its core idea originating from the diffusion process in non-equilibrium thermodynamics. The essence of the Diffusion Model comprises two key components: the forward noising process and the reverse generative process. The forward diffusion process is defined by a Markov chain, where the addition of noise at each step can be expressed as:
\begin{equation}
\label{eq:1}
q(\mathbf{x}_t \mid \mathbf{x}_{t-1}) = \mathcal{N} \left( \mathbf{x}_t; \sqrt{1-\beta_t}\, \mathbf{x}_{t-1}, \beta_t \mathbf{I} \right),
\end{equation}
where $\beta_t$ is the predefined noise variance, and $\mathbf{x}_t$ represents the noisy data at step $t$. After $T$ steps of diffusion, the data approaches a standard Gaussian distribution. The reverse generative process, on the other hand, uses a neural network to predict the noise or data, progressively reconstructing the sample.
\subsubsection{Hierarchical Prior Fusion}
The theoretical foundation of the Hierarchical Prior Fusion is built upon the framework of the standard VAE, with the core objective of modeling the complexity of data distributions through multi-layer latent variable structures. The standard VAE optimizes by maximizing the Evidence Lower Bound (ELBO), expressed as:
\begin{equation} 
\label{eq:2}
\mathcal{L}_{\text{VAE}} = \mathbb{E}_{q(z|x)}\left[\log p(x|z)\right] - \beta \cdot D_{\text{KL}}\left(q(z|x) \| p(z)\right),
\end{equation}
where $z$ denotes the latent variable, and $\beta$ is the weighting coefficient for the KL divergence term. However, the single-layer latent variable structure of standard VAEs struggles to effectively capture multi-scale features of data. To address this, Hierarchical VAEs introduce multi-layer latent variables $ z_1, z_2, \dots, z_L $ to model the data distribution in stages:
\begin{equation} 
\label{eq:2}
p_\theta(x) = \int \prod_{l=1}^L p_\theta(z_l | z_{l+1}) \cdot p_\theta(x | z_1) \, dz_1 \dots dz_L,
\end{equation}
where each layer $ z_l $ corresponds to feature representations at different abstraction levels. To optimize this structure, Hierarchical VAEs use a progressive training strategy: initially, higher-layer latent variables $ z_{l \geq 2} $ are frozen, and only $ z_1 $ is optimized for local features; as training progresses, higher layers are unfrozen and semantic constraints are incorporated, enabling a fusion from low-level to high-level semantics.

\subsubsection{Directed Acyclic Graph}
DAG is a fundamental structure in graph theory, consisting of a finite set of vertices and a set of directed edges connecting these vertices. Formally, a DAG is defined as a pair $G=(V,E)$ , where $V$ is a finite set of vertices, $E \subseteq V \times V$ is a set of directed edges, such that no directed cycles exist. The absence of cycles ensures that there is no path in the graph that starts and ends at the same vertex. In other words, for any vertex $v \subseteq V$, there does not exist a sequence of edges leading from $v$ back to itself.

\subsection{Causal Matrix}
Before training begins, the data is first processed through the Causal Extraction module. The causal relationships among variables in tabular data are often unknown and it is difficult to obtain manually annotated causal graphs, especially in the presence of high-dimensional or complex nonlinear relationships. Therefore, this paper adopts the notears framework to automatically learn causal structures among variables from observational data. By transforming the traditional combinatorial optimization problem into a differentiable continuous optimization problem, the notears framework can effectively discover causal graphs that satisfy the DAG constraint without relying on expert prior knowledge.

Given the existence of complex nonlinear relationships in tabular data, we employ the nonlinear extension of notears based on a MLP, which uses a neural network to model the relationships between each variable and the others. The objective function is formulated as:
\begin{equation}
\min_{\omega} \frac{1}{2n} \left\| X - \mathcal{F}_\omega(X) \right\|_F^2 
+ \alpha \left\| \omega \right\|_1 
+ \beta \left\| \omega \right\|_2^2
\end{equation}
The learnable parameters in the model are denoted as $ \omega $, and the nonlinear mapping is modeled as $ \mathcal{F}_\omega(\cdot) $. Meanwhile, $ \alpha $ and $ \beta $ are adopted as regularization coefficients for sparsity constraint and weight decay, respectively. To ensure that the learned causal structure is semantically valid, interpretable, and consistent with the nature of causal relationships among variables in the real world, the model introduces a DAG constraint:
\begin{equation}
    h(\omega) = \mathrm{Tr}\left(e^{\mathcal{H}(\omega)}\right) - d = 0,
\end{equation}
where $ \mathcal{H}(\omega) $ is a matrix derived from the model parameters $ \omega $, $ \mathrm{Tr}(\cdot) $ denotes the trace of a matrix, $ e^{H(\theta)} $ represents the matrix exponential, and $ d $ is the dimensionality of the variable space. Finally, we obtain a weight matrix $ A \in \mathbb{R}^{d \times d} $ that represents the causal strength among variables. By setting a threshold $ \tau $ (e.g., 0.3), this matrix is transformed into a binary causal graph $ G \in \{0, 1\}^{d \times d} $, defined as:

\begin{equation}
    G_{ij} = 
    \begin{cases}
        1, & \text{if } |A_{ij}| > \tau \\
        0, & \text{otherwise}
    \end{cases}
\end{equation}

Each row in this causal matrix indicates the set of parent nodes for the corresponding variable. This structured causal graph is then introduced as prior knowledge in subsequent modeling stages, enhancing both the interpretability and performance of the generative model.

\subsection{Architecture}
\label{Architecture}
In this section, we present the overall framework and training mechanism of CausalDiffTab, which integrates causal discovery with a diffusion model to jointly model numerical and categorical features in tabular data. The causal matrix is then obtained via post-processing. During training, this causal matrix is matched with the noise matrix generated by the denoising network through causal pair matching, producing the loss terms $\mathcal{L}_{\text{causal}}^{\text{c}}$ and $\mathcal{L}_{\text{causal}}^{\text{n}}$. These terms serve as regularization to guide the model in perceiving causal relationships across different categories.

For numerical features, we model the forward process $X^{n}$ using a stochastic differential equation (SDE) of the form:
\begin{equation}
\label{eq:3}
dX_t = f(X_t, t)\,dt + g(t)\,dW_t,
\end{equation}
where $f(\cdot, t): \mathbb{R}^{M_n} \to \mathbb{R}^{M_n}$ denotes the drift coefficient, $g(\cdot):\mathbb{R} \to \mathbb{R}$ represents the diffusion coefficient, and $W_t$ is a standard Wiener process \cite{song2021scorebased, karras2022elucidating}. The forward equation for numerical features is given by:
\begin{equation}
\label{eq:4}
x_{\text{n}}^t = x_{\text{n}}^0 + \sigma_{\text{n}}(t)\varepsilon, \quad \varepsilon \sim \mathcal{N}(0, I_{M_{\text{n}}}),
\end{equation}
and the reversal can then be formulated accordingly as:
\begin{equation}
\label{eq:5}
dx_{\text{n}} = -\left[\frac{d}{dt}\sigma_{\text{n}}(t)\right]\sigma_{\text{n}}(t)\nabla_x \log p_t(x_{\text{n}})dt,
\end{equation}
we use $\mu_{\text{n}}$ to denote the numerical component of the output of denoising network. It is trained by minimizing the denoising loss:
\begin{equation}
\label{eq:6}
L_{\text{n}}(\theta, \rho) = \mathbb{E}_{x_0 \sim p(x_0)} \mathbb{E}_{t \sim U[0,1]} \mathbb{E}_{\varepsilon \sim \mathcal{N}(0, I)} \left\| \mu_{\text{n}}^\theta(x_t, t) - \varepsilon \right\|_2^2.
\end{equation}

For categorical features, we first apply one-hot encoding to them. The forward diffusion process is defined as smoothly interpolating between the data distribution $\text{c}(\cdot; x)$ and the target distribution $\text{c}(\cdot; m)$, where all probability mass is concentrated on the [MASK] state:
\begin{equation}
\label{eq:7}
q(x_t | x_0) = \text{c}(x_t; \alpha_t x_0 + (1 - \alpha_t)m),
\end{equation}
$\alpha_t \in [0, 1]$ be a strictly decreasing function of $t$, with $\alpha_0 \approx 1$ and $\alpha_1 \approx 0$. It represents the probability for the real data $x_0$ to be masked at time step $t$. where $\sigma_{\text{c}}(t) : [0, 1] \to \mathbb{R}^+$ is a strictly increasing function. Such forward process entails the step transition probabilities: $q(x_t | x_s) = \text{c}(x_t; \alpha_{t|s} x_s + (1 - \alpha_{t|s})\mathbf{m})$. where $\alpha_{t|s} = \frac{\alpha_t}{\alpha_s}$. Under the hood, this transition means that at each diffusion step, the data will be perturbed to the [MASK] state with a probability of $(1 - \alpha_{t|s})$, and remains there until $t = 1$ if perturbed. The diffusion model $\mu_\theta$ aims to progressively uncover each column from the ``masked'' state.
\begin{equation}
\label{eq:8}
q(\mathbf{X}_s | \mathbf{X}_t, \mathbf{X}_0) =
\begin{cases}
\text{c}(\mathbf{x}_s; \mathbf{x}_t), & \mathbf{x}_t \neq \mathbf{m}, \\
\text{c}\left(\mathbf{x}_s; \frac{(1-\alpha_s)\mathbf{m} + (\alpha_s - \alpha_t)\mathbf{x}_0}{1 - \alpha_t}\right), & \mathbf{x}_t = \mathbf{m}.
\end{cases}
\end{equation}
increasing the discretization resolution can help approximate a tighter ELBO. Therefore, we optimize the likelihood bound $\mathcal{L}_{\text{c}}$ under the continuous-time limit, $\alpha'_t$ is the first order derivative of $\alpha_t$:
\begin{equation}
\label{eq:9}
\mathcal{L}_{\text{c}}(\theta, k) = \mathbb{E}_q \int_{t=0}^{t=1} \frac{\alpha_t'}{1 - \alpha_t} 1_{\{x_t = m\}} \log \langle \boldsymbol{\mu}_{\theta}^{\text{c}}(\mathbf{x}_t, t), \mathbf{x}_0^{\text{c}} \rangle dt,
\end{equation}
 
First, the encoded data is fed into the causal extraction module, as introduced in Section 3.2, resulting in a causal matrix that captures the causal relationships within the data. Second, causal loss is obtained by matching the model's predicted noise values with the causal matrix through causal pair alignment. Here, we compute the causal loss by calculating the outer product. By performing an outer product operation on the noise matrix generated by the model, we obtain a matrix that reflects the interaction strength between each pair of features. This representation not only preserves the information of the original features but also reveals potential causal directions. Therefore, by combining this with a pre-extracted causal matrix, we can measure the inconsistency between the predictions and the known causal structure—specifically, retaining correlations aligned with the allowed causal directions while suppressing those that violate causality. This helps guide the model to learn data generation mechanisms that conform to causal principles during training, and also effectively enhances the model's robustness. Finally, to better model different types of features, we separately compute the causal losses for numerical and categorical features. Specifically, the causal loss function is defined as follows:
\begin{equation}
\label{eq:10}
\mathcal{L}_{\text{causal}} = \lambda \cdot \mathbb{E}_{\text{batch}} \left[ \frac{1}{|\mathcal{M}|} \sum_{(i,j) \in \mathcal{M}} (\hat{\epsilon}_i \cdot \hat{\epsilon}_j) \right],
\end{equation}
where $\lambda$ is the regularization weight, $\hat{\boldsymbol{\epsilon}} = [\hat{\epsilon}_1, \hat{\epsilon}_2, \dots, \hat{\epsilon}_d]$ represents the model's predicted noise values (with $d$ feature dimensions), $\mathcal{M}$ denotes the causal mask, and $|\mathcal{M}|$ indicates the total number of non-zero elements in the mask. The total loss function integrates the diffusion model's base loss with the causal regularization term, defined as:
\begin{equation}
\label{eq:11}
\mathcal{L}_{\text{total}} = \mathcal{L}_{\text{c}}+ L_{\text{n}} + \lambda 
 (\mathcal{L}_{\text{causal}}^{\text{c}}+\mathcal{L}_{\text{causal}}^{\text{n}}).
\end{equation}

\begin{algorithm}[t]
\caption{Hybrid Adaptive Causal Regularization}
\label{HACR}
\label{alg:hybrid_loss}
\begin{algorithmic}[1]
\REQUIRE $\bm{\hat{x}}$ (predictions), $\bm{x}_t$ (noisy data), 
         $\bm{\sigma}$ (noise scale), $b_{\text{c}}$ (categorical flag)
\ENSURE Causal regularization loss

\IF{categorical features}
    \STATE $\bm{p} \leftarrow \text{Softmax}(\bm{\hat{x}})$
    \STATE $\bm{G} \leftarrow \bm{p}^\top \bm{p}$ \COMMENT{Categorical gradient outer product}
\ELSE
    \STATE $\bm{G} \leftarrow \left(\frac{\bm{x}_t-\bm{\hat{x}}}{\bm{\sigma}}\right)^\top 
            \left(\frac{\bm{x}_t-\bm{\hat{x}}}{\bm{\sigma}}\right)$ \COMMENT{Numerical gradient outer product}
\ENDIF

\STATE $\bm{M} \leftarrow \text{CausalMask}(\bm{G})$ \COMMENT{Apply causal masking}
\STATE $L_{\text{base}} \leftarrow \text{Mean}(\bm{G} \odot \bm{M})$ \COMMENT{Base consistency loss}

\STATE $\Delta L \leftarrow |L_{\text{base}} - \text{EMA}(L_{\text{base}})|$ \COMMENT{Loss fluctuation}
\STATE $\sigma_{\mu} \leftarrow \text{Mean}(\bm{\sigma})$ \COMMENT{Noise level}
\STATE $w \leftarrow \frac{w_{\text{max}}}{2}\left(e^{-\Delta L} + \frac{1}{1+\sigma_{\mu}}\right)$ \COMMENT{Hybrid weight}

\STATE Update $\text{EMA}(L_{\text{base}}) \leftarrow \alpha L_{\text{base}} + (1-\alpha)\text{EMA}(L_{\text{base}})$

\RETURN $L_{\text{causal}} = L_{\text{base}} \cdot w$
\end{algorithmic}
\end{algorithm}

\begin{table*}[htbp]
    \centering
    \tabcolsep=0.2cm
    \caption{Overview of dataset characteristics. The column "\# Num" indicates the count of numerical features, while "\# Cat" represents the count of categorical features. Additionally, "\# Max Cat" specifies the maximum number of categories found within any single categorical feature in the dataset.}
    \label{tab:datasets}
    \begin{tabular}{lccccccc}
        \toprule[0.8pt]
        \textbf{Dataset} & \textbf{Adult} & \textbf{Default} & \textbf{Shoppers} & \textbf{Magic} & \textbf{Beijing} & \textbf{News} & \textbf{Diabetes}\\
        \midrule 
        \textbf{\# Rows} & $48,842$ & $30,000$ & $12,330$ & $19,019$ & $43,824$ & $39,644$& $101766$  \\
        \textbf{\# Num} & $6$ & $14$ & $10$ & $10$ & $7$ & $46$& $9$\\
        \textbf{\# Cat} & $9$ & $11$ & $8$ & $1$ & $5$ & $2$ & $27$\\
        \textbf{\# Max Cat} & $42$ & $11$ & $20$ & $2$ & $31$ & $7$& $716$ \\
        \textbf{\# Train} & $28,943$ & $24,000$ & $9,864$ & $15,215$ & $35,058$ & $31,714$ & $61059$\\
        \textbf{\# Validation} & $3,618$ & $3,000$ & $1,233$ & $1,902$ & $4,383$ & $3,965$ & $20353$\\
        \textbf{\# Test} & $16,281$ & $3,000$ & $1,233$ & $1,902$ & $4,383$ & $3,965$ & $20354$\\
        \textbf{Task} & Classification & Classification & Classification & Classification & Regression & Regression & Classification\\
        \bottomrule[1.0pt] 
    \end{tabular}
\end{table*}
\subsection{Hybrid Adaptive Causal Regularization}
The integration of causal loss with adaptive weighting mechanisms in diffusion models originates from the principle of Hierarchical Prior Fusion in generative models. Analogous to hierarchical variational inference in VAE, this theory emphasizes improving generation quality by injecting multi-level prior knowledge in stages: during the initial training phase, the model prioritizes learning low-level local features (e.g., texture, edges) and later integrates high-level regularization (e.g., causal relationships and physical laws). Specifically, our hybrid adaptive causal regularization jointly considers the loss fluctuation $\Delta L$ at each step and the noise level $\sigma_{\text{mean}}$ to control the weighting of causal regularization.
\begin{equation}
\label{eq:4}
w_{\text{hybrid}} = w_{\text{max}} \cdot \frac{1}{2} \left( e^{-|\Delta L|} + \frac{1}{1+\sigma_{\text{mean}}} \right),
\end{equation}
when $\sigma_{\text{mean}} \to 0$ (low noise) and $\Delta L \to 0$ (stable training), $w_{\text{hybrid}} \to w_{\text{max}}$, applying strong regularization. Conversely, when $\sigma_{\text{mean}} \to \infty$ or $\Delta L \to \infty$ (violent oscillations), $w_{\text{hybrid}} \to 0$, avoiding interference with model learning. During the forward process of diffusion models, early denoising steps correspond to high-noise microscopic state spaces. Imposing strong causal regularization directly at this stage may lead to erroneous causal associations (e.g., spurious correlations) due to insufficient decoupling of low-level features. As the denoising progresses and the feature space becomes clearer, gradually increasing the weight of causal loss naturally embeds causal priors as high-level semantic Regularization into the generative process. This strategy also enhances the model’s noise robustness. Assuming the hybrid adaptive noise factor $\frac{1}{1+\sigma_{\text{mean}}}$ (a monotonically decreasing function of $\sigma_{\text{mean}}$), we can derive:
\begin{equation}
\label{eq:5}
\frac{\partial}{\partial \sigma_{\text{mean}}} \left( \frac{1}{1+\sigma_{\text{mean}}} \right) = 
-\frac{1}{(1+\sigma_{\text{mean}})^2} < 0,
\end{equation}
a higher noise level leads to a lower regularization weight, which prevents overfitting to the noise. This strategy not only avoids optimization instability caused by competing multi-task objectives in early training stages but also ultimately achieves a balance between generation quality and causal consistency. To better understand the hybrid adaptive causal regularization, we summarize the pseudocode in \mbox{Algorithm \ref{HACR}}.

\section{Experiments}
We evaluate CausalDiffTab by comparing it with various generative models across multiple datasets and metrics, ranging from data fidelity and privacy to downstream task performance. Additionally, we conduct ablation studies to investigate the effectiveness of each component of CausalDiffTab.
\subsection{Experimental Settings}
We use PyTorch \cite{10.5555/3454287.3455008} to implement all the algorithms. All the experiments are conducted on NVIDIA 3090 GPU.
We adopt the same evaluation methodology and hyperparameters used in prior approaches \cite{shi2025tabdiff}, assessing the quality of synthetic data using seven distinct metrics divided into three groups: 1) Fidelity: Shape, trend, $\alpha$-precision, $\beta$-recall, and detection measures evaluate how effectively the synthetic data faithfully reconstructs the distribution of the real data; 2) Downstream Tasks: Machine Learning Efficacy (MLE) demonstrate the potential of the model to enhance downstream tasks; 3) Privacy: The Distance to Closest Record (DCR) score assesses the level of privacy protection by measuring the degree of similarity between the synthetic data and the training data. All reported experiment results are the average of 20 random sampled synthetic data generated by the best-validated models.

\subsection{Datasets}
Our evaluation spans seven real-world structured datasets$ \footnote{https://archive.ics.uci.edu/datasets} $: Adult, Default, Shopper, Magic, Fault, Beijing, News and Diabetes. Each dataset contains a combination of numerical and categorical features. These datasets are further categorized by their native machine learning objectives, falling under either classification or regression tasks. Comprehensive descriptions of their characteristics, including attribute distributions and task definitions, are documented in \mbox{Table \ref{tab:datasets}}.
\begin{table*}[htbp]
    \centering
    \tabcolsep=0.3cm
    \caption{Performance comparison of different methods based on shape similarity error rates (\%). Lower error rates indicate superior performance. \textcolor{blue}{Bold Face} highlights the best score for each dataset. OOM stands for "Out Of Memory."}
    \label{tab:shape}
    \begin{tabular}{lccccccc|c}
        \toprule[0.8pt]
        \textbf{Method} & \textbf{Adult} & \textbf{Default} & \textbf{Shoppers} & \textbf{Magic} & \textbf{Beijing} & \textbf{News} & \textbf{Diabetes}& \textbf{Average} \\
        \midrule
        CTGAN    & $16.84${\tiny$\pm$ $0.03$} & $16.83${\tiny$\pm$$0.04$} & $21.15${\tiny$\pm0.10$} & $9.81${\tiny$\pm0.08$}  &$21.39${\tiny$\pm0.05$} &   $16.09${\tiny$\pm 0.02$}&   $9.82${\tiny$\pm 0.08$} &  $15.99$  \\
        TVAE     & $14.22${\tiny$\pm0.08$} & $10.17${\tiny$\pm$$0.05$} & $24.51${\tiny$\pm0.06$} & $8.25${\tiny$\pm0.06$}  &  $19.16${\tiny$\pm0.06$}  &  $16.62${\tiny$\pm0.03$} &   $18.86${\tiny$\pm 0.13$}& $15.97$   \\
        GOGGLE   & $16.97${\tiny$\pm$$0.00$} & $17.02${\tiny$\pm$$0.00$} & $22.33${\tiny$\pm$$0.00$} & $1.90${\tiny$\pm$$0.00$}  & $16.93 ${\tiny$\pm$$0.00$} &   $25.32${\tiny$\pm$$0.00$} & $24.92${\tiny$\pm$$0.00$}&  $17.91$ \\
        GReaT     & $12.12${\tiny$\pm$$0.04$} & $19.94${\tiny$\pm$$0.06$}  & $14.51${\tiny$\pm0.12$}  &  $16.16${\tiny$\pm0.09$}   & $8.25${\tiny$\pm0.12$}  &  OOM & OOM& $14.20$ \\
        STaSy    & $11.29${\tiny$\pm0.06$} & $5.77${\tiny$\pm0.06$} & $9.37${\tiny$\pm0.09$} & $6.29${\tiny$\pm0.13$}   & $6.71${\tiny$\pm0.03$}  &  $6.89${\tiny$\pm0.03$} & OOM&  $7.72$ \\
         CoDi  & $21.38${\tiny$\pm0.06$}  & $15.77${\tiny$\pm$ $0.07$}  & $31.84${\tiny$\pm0.05$}  & $11.56${\tiny$\pm0.26$}  & $16.94${\tiny$\pm0.02$} &    $32.27${\tiny$\pm0.04$}&   $21.13${\tiny$\pm 0.25$} & $21.55$  \\
        TabDDPM & $1.75${\tiny$\pm0.03$}  & $1.57${\tiny$\pm$ $0.08$}  & $2.72${\tiny$\pm0.13$}  & $1.01${\tiny$\pm0.09$}   & $1.30${\tiny$\pm0.03$}  &  $78.75${\tiny$\pm0.01$} &   $31.44${\tiny$\pm 0.05$}& $16.93$ \\
        TABDIFF & ${0.78}${\tiny${\pm0.04}$}& ${1.29}${\tiny${\pm0.09}$}& ${1.47}${\tiny${\pm0.11}$}& ${0.82}${\tiny${\pm0.06}$}& ${1.07}${\tiny${\pm0.06}$}&${2.73}${\tiny${\pm0.02}$}&${1.16}${\tiny${\pm0.02}$} & $1.33 $      \\
        \midrule
        Ours    & \textcolor{blue}{$\textbf{0.66}${\tiny${\pm0.06}$}}& \textcolor{blue}{$\textbf{1.15}${\tiny${\pm0.06}$}}&\textcolor{blue}{$\textbf{1.23}${\tiny${\pm0.11}$}}& \textcolor{blue}{$\textbf{0.79}${\tiny${\pm0.09}$}}&\textcolor{blue}{$\textbf{1.00}${\tiny${\pm0.05}$}}&\textcolor{blue}{$\textbf{2.28}${\tiny${\pm0.03}$}}&\textcolor{blue}{$\textbf{1.10}${\tiny$\pm 0.03$}}&\textcolor{blue}{ \textbf{1.17}}      \\
        Improv. &${18.2\% \downarrow} $ & ${12.2\% \downarrow} $ & ${19.5\% \downarrow}$  & ${3.8\% \downarrow} $  & ${7\% \downarrow}$  &  ${19.7\% \downarrow}$ & ${5.4\% \downarrow}$& ${13.7\% \downarrow}$  \\
        \bottomrule[1.0pt]
    \end{tabular}
\end{table*}

\begin{table*}[htbp]
    \centering
    \tabcolsep=0.3cm
    \caption{Performance comparison of different methods based on trend error rates (\%). Lower error rates reflect higher performance.}
    \label{tab:trend}
    \begin{tabular}{lccccccc|c}
        \toprule[0.8pt]
        \textbf{Method} & \textbf{Adult} & \textbf{Default} & \textbf{Shoppers} & \textbf{Magic}  & \textbf{Beijing} & \textbf{News} & \textbf{Diabetes}& \textbf{Average}  \\
        \midrule 
        CTGAN    & $20.23${\tiny$\pm1.20$} & $26.95${\tiny$\pm0.93$} & $13.08${\tiny$\pm0.16$} & $7.00${\tiny$\pm0.19$} &  $22.95${\tiny$\pm0.08$}  &  $5.37${\tiny$\pm0.05$}&  $18.95${\tiny$\pm0.34$} & $16.36$     \\
        TVAE     & $14.15${\tiny$\pm0.88$} & $19.50${\tiny$\pm$$0.95$} & $18.67${\tiny$\pm0.38$} &  $5.82${\tiny$\pm0.49$}  &  $18.01${\tiny$\pm0.08$}  & $6.17${\tiny$\pm0.09$} &  $32.74${\tiny$\pm0.26$}&$16.44$   \\
        GOGGLE   & $45.29${\tiny$\pm$$0.00$} & $21.94${\tiny$\pm$$0.00$} & $23.90${\tiny$\pm$$0.00$} & $9.47${\tiny$\pm$$0.00$}  & $45.94${\tiny$\pm$$0.00$}  & $23.19${\tiny$\pm$$0.00$} & $27.56${\tiny$\pm$$0.00$}& $28.18$   \\
        GReaT    & $17.59${\tiny$\pm0.22$} & $70.02${\tiny$\pm$$0.12$} & $45.16${\tiny$\pm0.18$} & $10.23${\tiny$\pm0.40$}  & $59.60${\tiny$\pm0.55$} & OOM &OOM&$44.24$  \\
        STaSy    & $14.51${\tiny$\pm0.25$} & $5.96${\tiny$\pm$$0.26$}  & $8.49${\tiny$\pm0.15$} & $6.61${\tiny$\pm0.53$}  & $8.00${\tiny$\pm0.10$} & $3.07${\tiny$\pm0.04$} & OOM&$7.77$     \\
        CoDi  & $22.49${\tiny$\pm0.08$}  & $68.41${\tiny$\pm$$0.05$}  & $17.78${\tiny$\pm0.11$}  & $6.53${\tiny$\pm0.25$} & $7.07${\tiny$\pm0.15$} & $11.10${\tiny$\pm0.01$}& $29.21${\tiny$\pm0.12$} & $23.21$   \\ 
        TabDDPM & $3.01${\tiny$\pm0.25$}  & $4.89${\tiny$\pm0.10$}  & $6.61${\tiny$\pm0.16$} & $1.70${\tiny$\pm0.22$} & $2.71${\tiny$\pm0.09$} & $13.16${\tiny$\pm0.11$}& $51.54${\tiny$\pm0.05$} & $11.95$ \\
        TABDIFF &${1.66}${\tiny${\pm0.21}$}& ${4.80}${\tiny${\pm1.66}$}& ${1.93}${\tiny${\pm0.08}$}&\textcolor{blue}{$\textbf{0.88}${\tiny${\pm0.24}$}} & ${2.72}${\tiny${\pm0.09}$}&${1.99}${\tiny${\pm0.02}$}  &${2.81}${\tiny${\pm0.02}$}   & 2.40 \\
        \midrule
        Ours   &\textcolor{blue}{$\textbf{1.55}${\tiny${\pm0.17}$}}&\textcolor{blue}{ $\textbf{3.43}${\tiny${\pm0.75}$}}& \textcolor{blue}{$\textbf{1.82}${\tiny${\pm0.09}$}}& ${0.90}${\tiny${\pm0.24}$} & \textcolor{blue}{$\textbf{2.65}${\tiny${\pm0.103}$}} &\textcolor{blue}{$\textbf{1.70}${\tiny${\pm0.01}$}}&\textcolor{blue}{$\textbf{2.67}${\tiny${\pm0.08}$}}& \textcolor{blue}{\textbf{2.10}}       \\
        Improve. &  ${7.1\% \downarrow} $ & ${43.71\% \downarrow} $ & ${6.0\% \downarrow}$  & ${0\% \downarrow} $  & ${2.6\% \downarrow}$  &  ${17.1\% \downarrow}$ & ${5.2\% \downarrow}$ & ${14.3\% \downarrow}$ \\
		\bottomrule[1.0pt] 
    \end{tabular}
\end{table*}
\subsection{Baselines}
We compare MoR with nine popular synthetic tabular data generation that are categorized into four groups: 1) GAN-based method: CTGAN \cite{NEURIPS2019_254ed7d2}; 2) VAE-based methods: TVAE \cite{NEURIPS2019_254ed7d2}, GOGGLE \cite{liu2023goggle}; 3) Autoregressive Language Model: GReaT; 4) Diffusion-based methods: STaSy \cite{kim2023stasy}, CoDi \cite{pmlr-v202-lee23i}, TabDDPM \cite{10.5555/3618408.3619133}, 
and TABDIFF \cite{shi2025tabdiff}.
\begin{table*}[htbp]
    \centering
    \tabcolsep=0.3cm
    \caption{Performance comparison of different methods based on $\alpha$-Precision scores. Higher error rates reflect higher performance. }
    \label{tab:alpha}
        \begin{tabular}{lcccccc|cc}
            \toprule[0.8pt]
            {Methods} & \textbf{Adult} & \textbf{Default} & \textbf{Shoppers} & \textbf{Magic} & \textbf{Beijing} & \textbf{News} & \textbf{Average} & \textbf{Ranking} \\
            \midrule 
            CTGAN    & $77.74${\tiny$\pm0.15$}  & $62.08${\tiny$\pm0.08$} & $76.97${\tiny$\pm0.39$} & $86.90${\tiny$\pm0.22$} & $96.27${\tiny$\pm0.14$} & $96.96${\tiny$\pm0.17$} & $82.82$ & $6$ \\
            TVAE     & $98.17${\tiny$\pm0.17$}  & $85.57${\tiny$\pm0.34$} & $58.19${\tiny$\pm0.26$} & $86.19${\tiny$\pm0.48$} & $97.20${\tiny$\pm0.10$} & $86.41${\tiny$\pm0.17$} & $85.29$  & $5$ \\
            GOGGLE  & $50.68${\tiny$\pm$$0.00$}  & $68.89${\tiny$\pm$$0.00$} & $86.95${\tiny$\pm$$0.00$} & $90.88${\tiny$\pm$$0.00$} & $88.81${\tiny$\pm$$0.00$}& $86.41${\tiny$\pm$$0.00$} & $78.77$ & $8$ \\
            GReaT    & $55.79${\tiny$\pm0.03$}  & $85.90${\tiny$\pm0.17$}  & $78.88${\tiny$\pm0.13$} & $85.46${\tiny$\pm0.54$} & ${98.32}${\tiny$ {\pm0.22}$} & OOM & $80.87$  & $6$ \\
            STaSy    & $82.87${\tiny$\pm0.26$} & $90.48${\tiny$\pm0.11$} & $89.65${\tiny$\pm0.25$} & $86.56${\tiny$\pm0.19$} & $89.16${\tiny$\pm0.12$} & $94.76${\tiny$\pm0.33$}  & $88.91$ & $3$ \\
            CoDi & $77.58${\tiny$\pm0.45$} & $82.38${\tiny$\pm0.15$}  & $94.95${\tiny$\pm0.35$} & $85.01${\tiny$\pm0.36$} & ${98.13}${\tiny${\pm0.38}$} & $87.15${\tiny$\pm0.12$} & $87.53$ & $4$ \\
            TabDDPM  & $96.36${\tiny$\pm0.20$}  & $97.59${\tiny$\pm0.36$} & $88.55${\tiny$\pm0.68$} & $98.59${\tiny$\pm0.17$} & $97.93$\tiny${\pm0.30}$ & $0.00${\tiny$\pm0.00$} &  $79.84$  & $7$ \\
            TABDIFF & 
            ${97.99}$\tiny${\pm0.29}$  & 
            ${98.44}$\tiny${\pm0.36}$ & ${99.11}$\tiny$ {\pm0.26}$ & 
            $ {99.07}$\tiny$ {\pm0.35}$ & 
            ${98.06}$\tiny${\pm0.24}$ & 
            $ {90.40}$\tiny$ {\pm0.34}$ & 
            $ {97.18}$ & $2$ \\
            \midrule
            Ours& ${99.03}$\tiny${\pm0.27}$ &${98.57}$\tiny${\pm0.33}$& ${99.42}$\tiny${\pm0.18}$& ${99.40}$\tiny${\pm0.28}$& ${97.80}$\tiny${\pm0.47}$ & ${95.20}$\tiny${\pm0.38}$&{98.24} & 1\\
		\bottomrule[1.0pt] 
		\end{tabular}
\end{table*}

\begin{table*}[htbp]
    \centering
    \tabcolsep=0.3cm
    \caption{Performance comparison of different methods based on $\beta$-Recall scores. Higher error rates reflect higher performance. }
    \label{tab:beta}
        \begin{tabular}{lcccccc|cc}
            \toprule[0.8pt]
            {Methods} & \textbf{Adult} & \textbf{Default} & \textbf{Shoppers} & \textbf{Magic} & \textbf{Beijing} & \textbf{News} & \textbf{Average}  & \textbf{Ranking} \\
            \midrule 
            CTGAN    & $30.80${\tiny$\pm0.20$}  & $18.22${\tiny$\pm0.17$} & $31.80${\tiny$\pm0.350$} & $11.75${\tiny$\pm0.20$} & $34.80${\tiny$\pm0.10$} & $24.97${\tiny$\pm0.29$}  & $25.39$ & $8$ \\
            TVAE     & $38.87${\tiny$\pm0.31$}  & $23.13${\tiny$\pm0.11$} & $19.78${\tiny$\pm0.10$} & $32.44${\tiny$\pm0.35$} & $28.45${\tiny$\pm0.08$} & $29.66${\tiny$\pm0.21$}  & $28.72$  & $7$  \\
            GOGGLE  & $8.80${\tiny$\pm$$0.00$}  & $14.38${\tiny$\pm$$0.00$} & $9.79${\tiny$\pm$$0.00$} & $9.88${\tiny$\pm$$0.00$} & $19.87${\tiny$\pm$$0.00$} & $2.03${\tiny$\pm$$0.00$} & $10.79$ & $9$ \\
            GReaT    & ${49.12}${\tiny${\pm0.18}$}  & $42.04${\tiny$\pm0.19$}  & $44.90${\tiny$\pm0.17$} & $34.91${\tiny$\pm0.28$} & $43.34${\tiny$\pm0.31$} & OOM & $43.86$ & $3$ \\
            STaSy    & $29.21${\tiny$\pm0.34$} & $39.31${\tiny$\pm0.39$} & $37.24${\tiny$\pm0.45$} & ${53.97}${\tiny${\pm0.57}$} & $54.79${\tiny$\pm0.18$} & $39.42${\tiny$\pm0.32$} &$42.32$ & $4$ \\
            CoDi & $9.20${\tiny$\pm0.15$} & $19.94${\tiny$\pm0.22$}  & $20.82${\tiny$\pm0.23$} & $50.56${\tiny$\pm0.31$} & $52.19${\tiny$\pm0.12$} & $34.40${\tiny$\pm0.31$} &  $31.19$ & $6$ \\
            TabDDPM  & $47.05${\tiny$\pm0.25$}  & $47.83${\tiny$\pm0.35$} & $47.79${\tiny$\pm0.25$} & $ {48.46}${\tiny$ {\pm0.42}$} & ${56.92}$\tiny${\pm0.13}$ & $0.00${\tiny$\pm0.00$} & 
            $41.34$ & $5$  \\
            TABDIFF & 
            $ {51.64}$\tiny$ {\pm0.20}$  & 
            $ {52.47}$\tiny$ {\pm0.36}$ & 
            $ {48.26}$\tiny$ {\pm0.57}$ & 
            ${48.87}$\tiny${\pm0.41}$ & 
            $ {59.62}$\tiny$ {\pm0.19}$ & 
            ${35.52}$\tiny${\pm0.21}$ & 
            $ {49.40}$  & $2$ \\
            \midrule
            Ours & $ {51.31}$\tiny$ {\pm0.24}$ &$ {51.45}$\tiny$ {\pm0.40}$ &$ {49.43}$\tiny$ {\pm0.69}$ & $ {48.34}$\tiny$ {\pm0.48}$ & $ {59.55}$\tiny$ {\pm0.23}$ & $ {41.31}$\tiny$ {\pm0.21}$ &50.32 & 1\\
		\bottomrule[1.0pt] 
		\end{tabular}
\end{table*}

\subsection{Data Fidelity}
We first evaluate the shape and trend metrics. The shape metric measures the ability of synthetic data to capture the marginal density of each individual column, while the trend metric assesses its capability to replicate correlations between different columns in real data. Detailed results are presented in Table \ref{tab:shape} and Table \ref{tab:trend}. Analysis reveals that CausalDiffTab outperforms all baselines across six datasets in terms of shape metrics. It surpasses the current state-of-the-art method with an average improvement of 14.3\%, demonstrating its superior performance in maintaining marginal distributions for various attributes across datasets. In terms of trend metrics, CausalDiffTab also demonstrates excellent and robust performance. Notably, on the Default task, CausalDiffTab achieves a 43.71\% improvement in metrics. This indicates that our method is significantly more effective at capturing complex relationships between columns compared to previous approaches, showcasing its superior modeling capability.

Furthermore, we evaluated fidelity metrics including $\alpha$-Precision, $\beta$-Recall, and CS2T scores. CS2T reflects the difficulty of distinguishing synthetic data from real data. The results are shown in Table \ref{tab:CS2T}. CausalDiffTab performs well across various tasks, especially on the Adult task, where it achieves an improvement of 1.31\%. Across all results, the average improvement is 0.57\%. This demonstrates that the causal regularization enhance the causal consistency of the generated data, making it more aligned with the conditional probability distribution of the real data. $\alpha$-Precision measures the quality of common data characteristics, where higher scores indicate greater faithfulness of synthetic data to real data. $\beta$-Recall assesses the extent to which synthetic data covers the distribution of real data. Due to space limitations, detailed results are provided in the supplementary material. On average, CausalDiffTab outperforms other methods across all three metrics.

\subsection{Data Privacy}
As the amount of data required for training generative models increases, an increasing number of privacy concerns face potential leakage risks. Synthetic tabular data can serve as a privacy-preserving alternative for AI training. In this section, we evaluate the privacy-preserving capabilities of the model using the DCR scoring metric as the evaluation criterion \cite{shi2025tabdiff}. The core logic involves assessing the similarity between generated data and original data to determine whether the model might expose sensitive information. The results are shown in Table \ref{tab:privacy}. CausalDiffTab outperforms the latest baselines on most datasets, highlighting its strong capability for privacy protection.
\begin{table*}[htbp]
    \centering
    \tabcolsep=0.35cm
    \caption{Performance comparison of different methods based on detection score (C2ST) using logistic regression classifier. Higher scores reflect superior performance.}
    \label{tab:CS2T}
        \begin{tabular}{lccccccc|c}
            \toprule[0.8pt]
             \textbf{Method} & \textbf{Adult} & \textbf{Default} & \textbf{Shoppers} & \textbf{Magic}  & \textbf{Beijing} & \textbf{News}& \textbf{Diabetes} & \textbf{Average} \\
            \midrule
            CTGAN & $0.5949$ & $0.4875$ & $0.7488$ & $0.6728$ & $0.7531$ & $0.6947$ & $0.5593$& $0.6444$  \\
            TVAE & $0.6315$  & $0.6547$ & $0.2962$ & $0.7706$ & $0.8659$ & $0.4076$ & $0.0487$& $0.5250$  \\
            GOGGLE & $0.1114$  & $0.5163$ & $0.1418$ & $0.9526$ & $0.4779$ & $0.0745$& $0.0912$& $0.3380$  \\
            GReaT & $0.5376$  & $0.4710$ & $0.4285$ & $0.4326$ & $0.6893$ & OOM & OOM&$0.5118$ \\
            STaSy & $0.4054$  & $0.6814$ & $0.5482$ & $0.6939$ & $0.7922$ & $0.5287$ &OOM& $0.6083$  \\
            CoDi & $0.2077$  & $0.4595$ & $0.2784$ & $0.7206$ & $0.7177$ & $0.0201$ & $0.0008$& $0.3435$  \\
            TabDDPM & $0.9755$  & $0.9712$ & $0.8349$ & $ \textcolor{blue}{0.9998}$ & $0.9513$ & $0.0002$& $0.1980$ &$0.7044$  \\
            TABDIFF &  $ {0.9832}$  & ${0.9615}$ & $ \textcolor{blue}{0.9843}$ & ${0.9944}$ & $ {0.9775}$ & $ {0.9308}$& $0.9521$ & ${0.9691}$ \\
            \midrule 
            Ours & \textcolor{blue}{0.9963} & \textcolor{blue}{0.9727} & 0.9812 & 0.9979 & \textcolor{blue}{0.9776} & \textcolor{blue}{0.9417}&\textcolor{blue}{0.9564} &\textcolor{blue}{0.9748} \\
		\bottomrule[1.0pt] 
		\end{tabular}
\end{table*}
\begin{table*}[htbp]
    \centering
    \tabcolsep=0.45cm
    \caption{Performance comparison of different methods based on DCR score. The DCR score measures the likelihood that a generated data sample resembles the training set more closely than the test set. A score closer to 50\% is more preferable.}
    \label{tab:privacy}
    \renewcommand\arraystretch{1.2}
        \begin{tabular}{lccccc|c}
            \toprule[0.8pt]
             \textbf{Method} & \textbf{Adult} & \textbf{Default} & \textbf{Shoppers} & \textbf{Beijing} & \textbf{News}& \textbf{Average}\\
            \midrule 
            TabDDPM & $51.14$\%{\tiny$\pm0.18$}  & $52.15$\%{\tiny$\pm0.20$} & $63.23$\%{\tiny$\pm0.25$} & $80.11$\%{\tiny$\pm2.68$}  & $79.31$\% {\tiny$\pm0.29$} & 65.18\% \\
            TABDIFF & $53.38$\%{\tiny$\pm0.37$}& {$56.52$\%}{\tiny$\pm0.38$} & {$51.11$\%}{\tiny$\pm0.54$}& $54.52$\%{\tiny$\pm0.27$}  & $55.04$\% {\tiny$\pm0.36$} & 54.11\%\\
            \midrule
            Ours & $54.47$\%{\tiny$\pm0.27$}& $56.29$\%{\tiny$\pm0.43$}& $50.77$\%{\tiny$\pm0.66$}& $53.58$\%{\tiny$\pm0.37$}& $54.04$\%{\tiny$\pm0.32$} & 53.83\%\\
		\bottomrule[1.0pt] 
        \end{tabular}
\end{table*}
\begin{table*}[htbp]
    \centering
    \tabcolsep=0.42cm
    \caption{Evaluation of MLE (Machine Learning Efficiency). For classification tasks, the higher the AUC score, the better the synthetic data quality; for regression tasks, the lower the RMSE, the better the quality.}
    \label{tab:mle}
        \begin{tabular}{lccccccc}
            \toprule[0.8pt]
            \multirow{2}{*}{Methods} & {\textbf{Adult}} &{\textbf{Default}} & \textbf{Shoppers} & {\textbf{Magic}} &   {\textbf{Beijing}} & {\textbf{News}} & {\textbf{Diabetes}}\\
            \cmidrule{2-8} 
            & AUC $\uparrow$ & AUC $\uparrow$ &  AUC $\uparrow$ &  AUC $\uparrow$  & RMSE $\downarrow$ &  RMSE $\downarrow$&  AUC $\uparrow$\\
            \midrule 
            Real & $.927${\tiny$\pm.000$} & $.770${\tiny$\pm.005$} & $.926${\tiny$\pm.001$}  & $.946${\tiny$\pm.001$}  & $.423${\tiny$\pm.003$} & $.842${\tiny$\pm.002$}& $.704${\tiny$\pm.002$}\\
            \midrule
            CTGAN & $.886${\tiny$\pm.002$} &$.696${\tiny$\pm.005$} & $.875${\tiny$\pm.009$} & $.855${\tiny$\pm.006$}    & $.902${\tiny$\pm.019$} & $.880${\tiny$\pm.016$}& $.569${\tiny$\pm.004$}\\
            TVAE  & $.878${\tiny$\pm.004$} &$.724 ${\tiny$\pm.005$} & $.871${\tiny$\pm.006$} & $.887${\tiny$\pm.003$} & $.770${\tiny$\pm.011$} & $1.01${\tiny$\pm.016$} & $.594${\tiny$\pm.009$} \\
            GOGGLE & $.778${\tiny$\pm.012$} & $.584${\tiny$\pm.005$} & $.658${\tiny$\pm.052$} & $.654${\tiny$\pm.024$}  & $1.09${\tiny$\pm.025$} & $.877${\tiny$\pm.002$} & $.475${\tiny$\pm.008$} \\
            GReaT & $.913${\tiny$\pm.003$} &$.755${\tiny$\pm.006$} & $.902${\tiny$\pm.005$} & $.888${\tiny$\pm.008$}  & $.653${\tiny$\pm.013$} & OOM&OOM  \\
            STaSy  & $.906${\tiny$\pm.001$} & $.752${\tiny$\pm.006$} & $.914${\tiny$\pm.005$} & $.934${\tiny$\pm.003$}  & $.656${\tiny$\pm.014$} & $.871${\tiny$\pm.002$}&OOM\\ 
            CoDi & $.871${\tiny$\pm.006$} & $.525${\tiny$\pm.006$} & $.865${\tiny$\pm.006$} & $.932${\tiny$\pm.003$}   & $.818${\tiny$\pm.021$} & $1.21${\tiny$\pm.005$}& $.505${\tiny$\pm.004$} \\
            TabDDPM & $.907${\tiny$\pm.001$}  & $.758${\tiny$\pm.004$}& $.918${\tiny$\pm.005$} & $.935${\tiny$\pm.003$} & $.592${\tiny$\pm.011$}& $4.86${\tiny$\pm3.04$}& $.521${\tiny$\pm.008$} \\
            TABDIFF & 
            ${.912}${\tiny${\pm.001}$} & 
            $.758${\tiny${\pm.011}$} & ${.918}${\tiny${\pm.003}$} & ${.936}${\tiny${\pm.003}$} & $.565${\tiny$\pm.012$} & ${.866}${\tiny${\pm.021}$} & $.688${\tiny$\pm.016$} \\
            \midrule
            Ours &\textcolor{blue}{${.914}${\tiny${\pm.001}$}} &\textcolor{blue}{${.761}${\tiny${\pm.007}$} } &\textcolor{blue}{${.921}${\tiny${\pm.004}$}}  &\textcolor{blue}{${.937}${\tiny${\pm.003}$}}  &\textcolor{blue}{${.561}${\tiny${\pm.012}$}} &\textcolor{blue}{${.863}${\tiny${\pm.022}$}} &\textcolor{blue}{${.692}${\tiny${\pm.002}$}}\\
		\bottomrule[1.0pt] 
		\end{tabular}
\end{table*}

\begin{table*}[htbp]
    \centering
    \tabcolsep=0.3cm
    \caption{Two sets of ablation experiment results: 1) the ablation study of CausalDiffTab using linear causal regularization and nonlinear causal regularization, respectively; 2) the ablation study of CausalDiffTab using Fixed Causal Regularization (FCR) and Hybrid Adaptive Causal Regularization (HACR), respectively. \textcolor{blue}{Bold Face} highlights the best score for each dataset.}
    \label{Ablation}
    \renewcommand\arraystretch{1.2}
    \begin{tabular}{l|ccccccc}
        \toprule[0.8pt]
        \textbf{Method} & \textbf{Tasks} & \textbf{Shape}$\downarrow$ & \textbf{Trend}$\downarrow$ & \textbf{MLE}$\uparrow$  & \textbf{CS2T}$\uparrow$  &\textbf{$\alpha$-Precision}$\uparrow$ & \textbf{$\beta$-Recall}$\uparrow$ \\
        \midrule 
        \multirow{2}{*}{Ours+NonLinear+HACR}&\textbf{Default}&\textcolor{blue}{1.15}&\textcolor{blue}{3.43}&\textcolor{blue}{0.76}&\textcolor{blue}{97.27}&\textcolor{blue}{98.57}&51.45\\
        &\textbf{Shopper}&\textcolor{blue}{1.23}&\textcolor{blue}{1.82}&\textcolor{blue}{0.92}&98.12&\textcolor{blue}{99.42}&\textcolor{blue}{49.43}\\
        \midrule
        \multirow{2}{*}{Ours+Linear+HACR}&\textbf{Default}&1.32&4.90&0.76&96.49&98.25&\textcolor{blue}{51.49}\\
        &\textbf{Shopper}&1.39&1.99&0.92&\textcolor{blue}{99.10}&99.27&47.50\\
        \midrule
        \multirow{2}{*}{Ours+Nonlinear+FCR}&\textbf{Default}&1.21&5.34&0.76&97.00&98.54&51.45\\
        &\textbf{Shopper}&1.44&1.95&0.92&97.49&99.27&47.50\\
		\bottomrule[1.0pt] 
    \end{tabular}
\end{table*}
\subsection{Performance on Downstream Tasks}
A key advantage of high-quality synthetic tabular data lies in its ability to serve as a substitute for real dataset in model training. In this section, we evaluate CausalDiffTab's capability to support downstream task learning through MLE measurement. The experimental design strictly follows domain-standard protocols \cite{NEURIPS2021_b578f2a5,borisov2023language,shi2025tabdiff}: First, we train the CausalDiffTab generative model on the original real dataset. Subsequently, we construct a synthetic dataset of the same scale as the original data using this trained model, which is then employed to train either an XGBoost classifier or XGBoostRegressor. For classification tasks, we calculate AUC scores, while regression tasks are quantified using RMSE metrics to measure prediction deviations. Finally, we assess the performance differences between models trained on synthetic data and those trained on real dataset. 
The experimental results presented in Table \ref{tab:mle} demonstrate that CausalDiffTab consistently achieves the best performance. Although the improvements are marginal, the results closely approach those obtained using real dataset for training. This validates that our method effectively corrects logical inconsistencies in the data, enabling the synthetic dataset to better approximate the outcomes of real dataset.
\begin{figure*}[htbp]
    \centering
    \begin{subfigure}[b]{0.48\textwidth}
        \includegraphics[width=\linewidth]{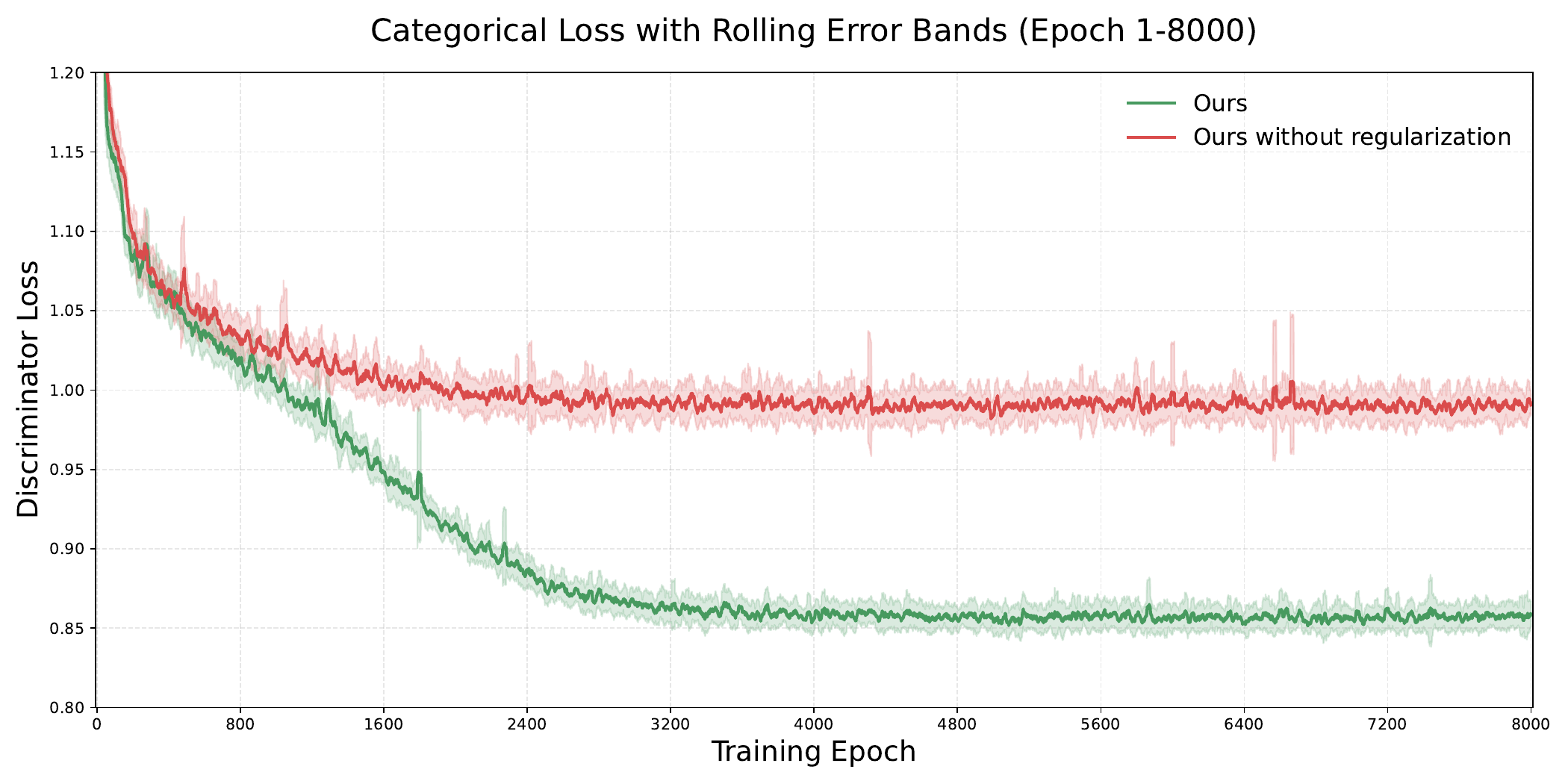}
    \end{subfigure}
    \hfill
    \begin{subfigure}[b]{0.48\textwidth}
        \includegraphics[width=\linewidth]{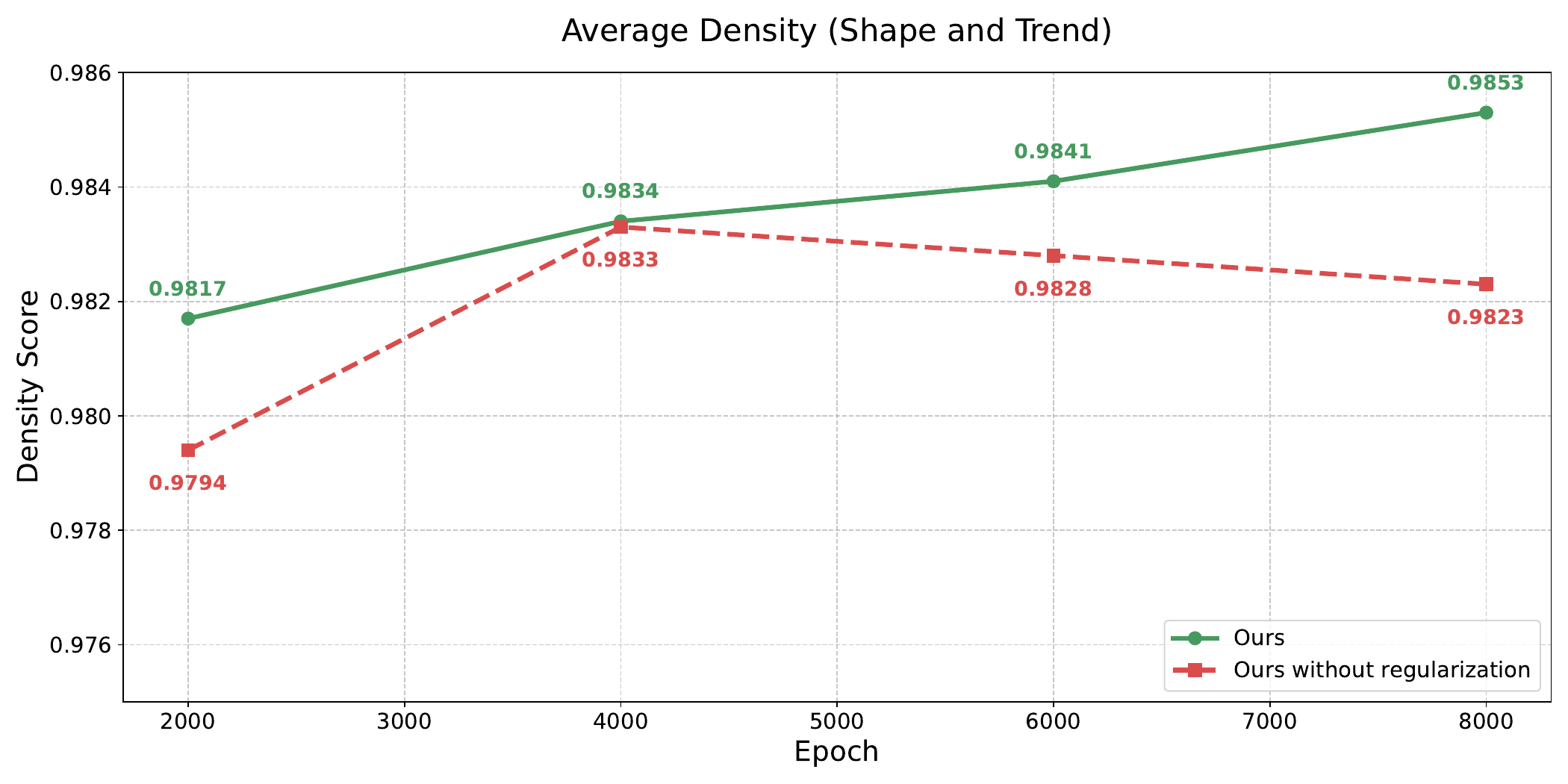}
    \end{subfigure}
    \vspace{-10pt}
    \caption{Comparison of training loss results and training evaluation (average of shape and trend) results with and without causal regularization on Shoppers tasks.}
    \label{fig:shop}
\end{figure*}
\begin{figure*}[htbp]
    \centering
    \begin{subfigure}[b]{0.48\textwidth}
        \includegraphics[width=\linewidth]{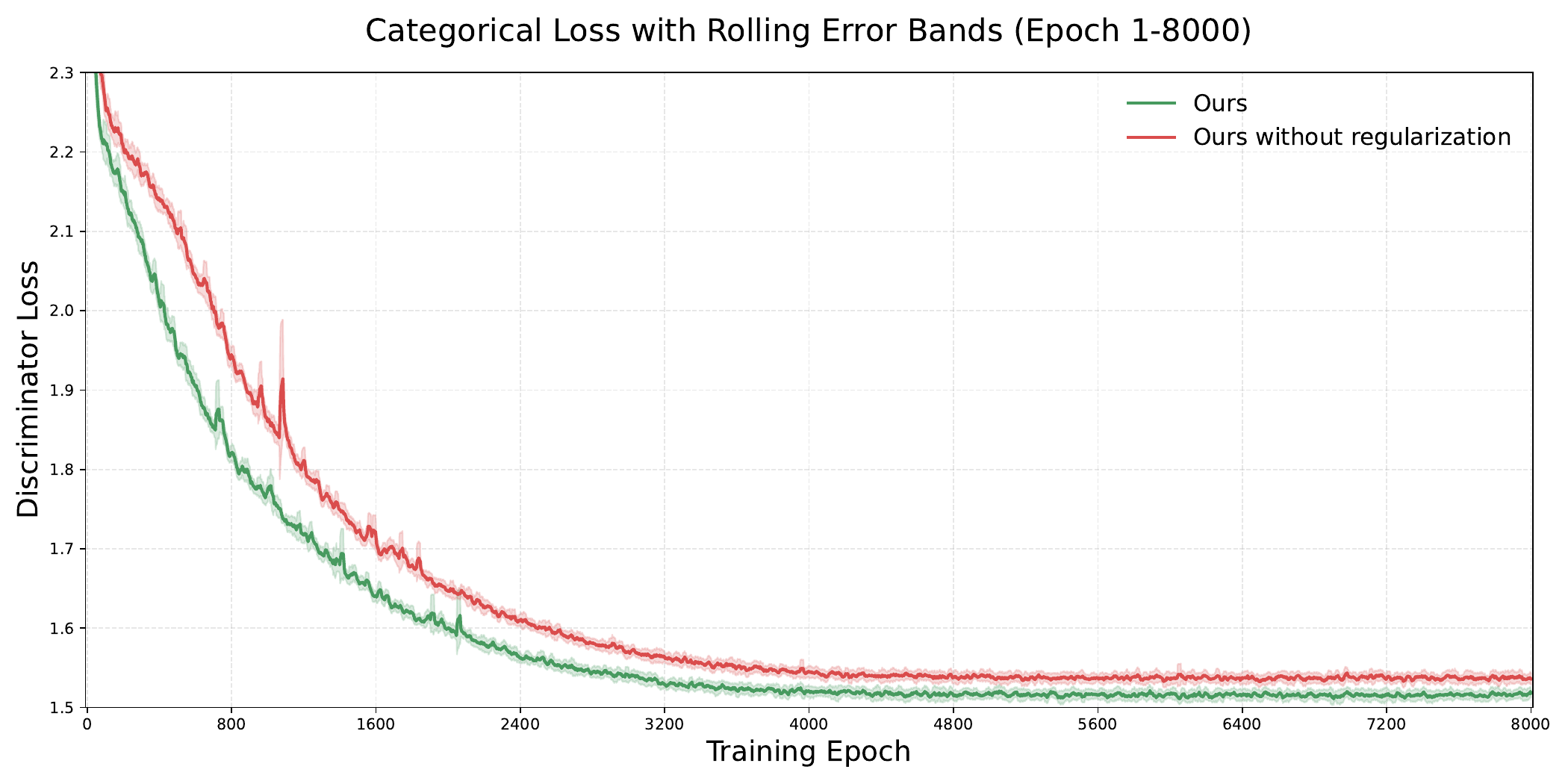}
    \end{subfigure}
    \hfill
    \begin{subfigure}[b]{0.48\textwidth}
        \includegraphics[width=\linewidth]{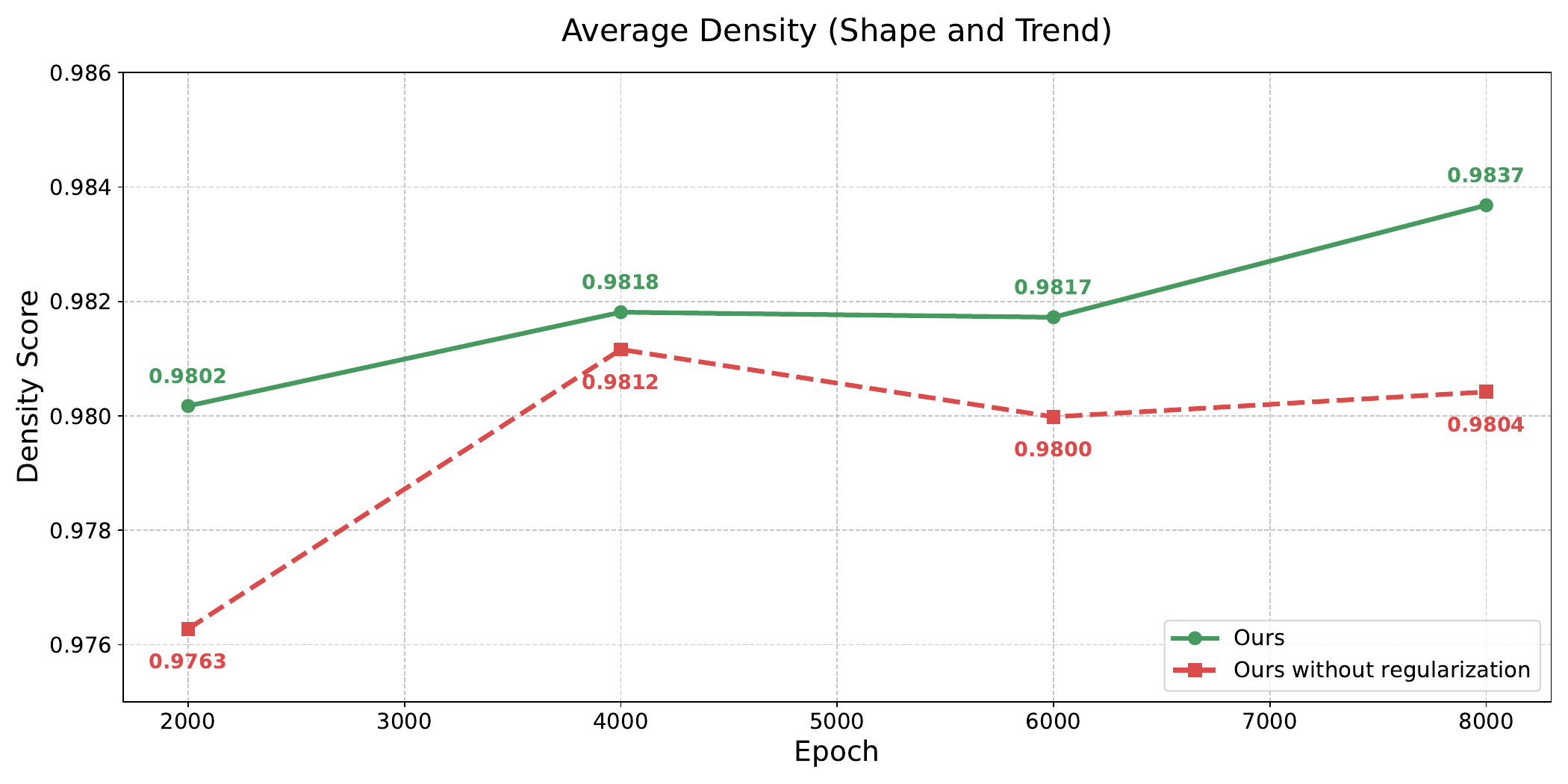}
    \end{subfigure}
    \vspace{-10pt}
    \caption{The results on Beijing tasks.}
    \label{fig:beijing}
\end{figure*}
\begin{figure}
  \includegraphics[width=\linewidth]{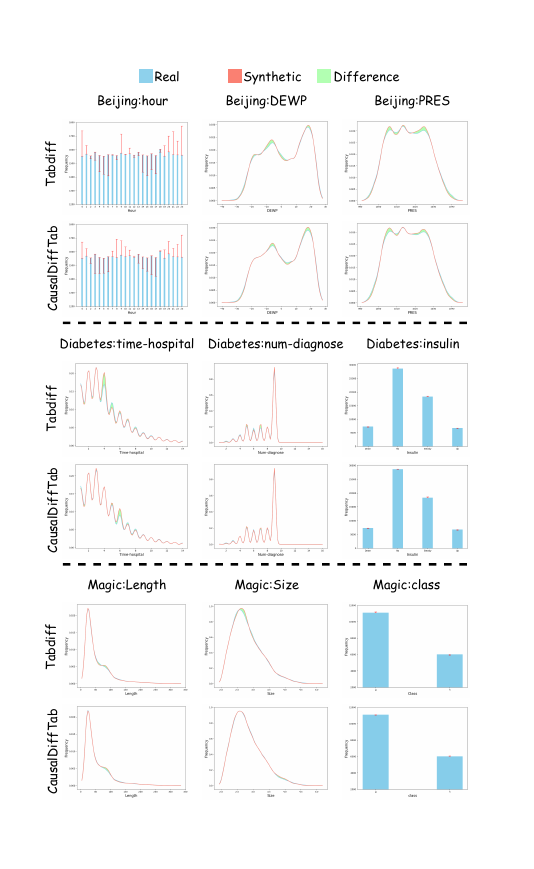}
  \caption{Visualization of marginal density comparisons between generated data and real data. Line plots represent numerical results, while bar charts represent categorical results.}
  \vspace{-10pt}
  \label{fig:vis}
\end{figure}
\subsection{Ablation Studies}
\subsubsection{Nonlinear Causal awareness}
We conduct ablation studies to evaluate the effectiveness of the nonlinear causal awareness proposed in Section \ref{Architecture}. We compare the results of incorporating nonlinear causal awareness with those of using linear causal awareness. Specifically, we adopt notears\_linear as the linear causal extractor and notears\_nolinear as the nonlinear causal extractor. Both methods use the same regularization strategies and number of iterations. The results are shown in Table \ref{Ablation}. The results show that introducing non-linear causal regularization in diffusion models allows for a more flexible capture of complex interactions between variables. Non-linear regularization effectively mitigate the over-simplification of data distributions caused by linear assumptions by dynamically adjusting the strength and direction of couplings. This flexibility enables the generation process to better align with the underlying non-linear causal mechanisms in real-world data, thereby significantly improving the causal plausibility of generated samples and their suitability for downstream tasks, while maintaining data diversity.
\subsubsection{Hybrid Adaptive Causal Regularization}
We further conduct a second ablation study to evaluate the effectiveness of the hybrid adaptive causal regularization proposed in Section \ref{HACR}. We compare the results obtained using hybrid adaptive causal regularization with those using fixed-value weights. The experiment employs notears\_nolinear as the nonlinear causal extractor. All other hyperparameters remain consistent. The results are presented in Table \ref{Ablation}. The results show that although adopting Fixed Causal Regularization has minimal impact on the $\alpha$-Precision and $\beta$-Recall metrics, it significantly leads to declines in other indicators, particularly the shape and trend metrics. This indicates that introducing causal regularization prematurely can impair the model's modeling and generation performance. In contrast, the hybrid adaptive causal regularization employed by CausalDiffTab effectively avoids optimization instability caused by competition among multi-task objectives during early training stages, ultimately achieving a balance between generation quality and causal consistency.

\section{Visualizations of Data}
We provide partial visualizations of the training process, as shown in Figures \ref{fig:shop} and \ref{fig:beijing}. These visualizations consistently indicate that the causal-aware module adopted by CausalDiffTab effectively accelerates model convergence and enhances training stability. We also present more detailed visualizations for some of the results, as shown in Figure \ref{fig:vis}. It can be observed that our method better fits the distribution of the original data.
\section{Conclusion}
In this paper, we propose CausalDiffTab, a diffusion-based generative model specifically designed for complex tabular data. CausalDiffTab incorporates a hybrid diffusion process to handle both numerical and categorical features in their native formats. To address potential causal misalignment issues, we perform DAG-based causal extraction on the original data, obtaining a causal matrix that captures the underlying causal relationships among variables. Guided by the principle of Hierarchical Prior Fusion, we develop a hybrid adaptive causal regularization technique that dynamically adjusts the weight of causal regularization based on loss fluctuations and noise levels. This approach effectively enhances the model's causal reasoning capabilities without sacrificing its generative performance. Through comprehensive evaluations on multiple real-world tabular datasets, our method achieves SOTAs, demonstrating its ability to produce high-quality, diverse synthetic data that faithfully supports downstream tasks.
\\ \\
\textbf{ CRediT authorship contribution statement}
\\
\par\textbf{Jia-Chen Zhang}: Writing – review \& editing, Writing – original draft, Conceptualization, Investigation, Methodology. 
\textbf{Zheng Zhou}: Investigation, Resources, Software. 
\textbf{Yu-Jie Xiong}: Funding acquisition, Investigation, Methodology, Project administration. \textbf{Chun-Ming Xia}: Project administration, Investigation, Writing—review and editing.
\textbf{Fei Dai}: Resources, Software, Supervision, Validation. \\ \\
\textbf{Declaration of competing interest}
\\
\par The authors declare that they have no known competing financial interests or personal relationships that could have appeared to influence the work reported in this paper.
\\ \\
\textbf{Acknowledgments}
\\
\par This work was supported in part by the Science and Technology Commission of Shanghai Municipality under Grant (21DZ2203100), in part by the National Natural Science Foundation of China under Grant (62006150), in part by Shanghai Local Capacity Enhancement project (21010501500) and in part by Science and Technology Innovation Action Plan of Shanghai Science and Technology Commission for social development project under Grant (21DZ1204900M).
\\ \\
\textbf{Data availability}
\\
\par The data used in this study are publicly available.

\bibliographystyle{cas-model2-names}

\bibliography{our}

\end{document}